\documentclass[letterpaper, 10 pt, conference]{ieeeconf}

\IEEEoverridecommandlockouts                              
\usepackage[T1]{fontenc}

\usepackage[noadjust]{cite}

\usepackage{filecontents}
\usepackage{amsmath}
\DeclareMathSymbol{:}{\mathord}{operators}{"3A}  
\usepackage{yhmath} 
\usepackage{mathtools} 
\usepackage{amssymb} 
\usepackage{xfrac} 
\usepackage{sidecap}
\usepackage{graphicx}
\usepackage{subcaption}
\usepackage{multirow} 
\usepackage{fancyhdr}
\usepackage{leftidx} 
\usepackage{cancel} 
\usepackage{caption}
\usepackage[super]{nth} 
\usepackage[normalem]{ulem} 
\usepackage{color}
\usepackage{xcolor} 
\usepackage{booktabs} 
\usepackage{siunitx} 
\usepackage{booktabs} 
\usepackage{colortbl} 
\usepackage{textcomp} 
\usepackage{tikz}
\usepackage[noadjust]{cite} 
\usepackage{xcolor}
\usepackage{wasysym}  
\DeclareMathOperator*{\argmin}{arg\,min} 
\newcommand{\RNum}[1]{\uppercase\expandafter{\romannumeral #1\relax}} 

\newcommand{\realfield}[1]{\hbox{I \kern -.5em R}^{#1}}
\newcommand {\mb}[1]{\mathbf{#1}}
\newcommand {\bs}[1]{\boldsymbol{#1}}
\newcommand{\uvec}[1]{\hat{\mathbf{#1}}}

\newcommand{\T}{^{\mathrm{T}}}  

\newcommand*\circled[1]{\tikz[baseline=(char.base)]{\node[circle,minimum size=7pt,draw=black,inner sep=0.5pt](char){\scriptsize #1};}}

\usepackage{enumerate}

\usepackage{color} 

\usepackage[textsize=scriptsize, bordercolor=white,backgroundcolor=gray!30,linecolor=black,colorinlistoftodos]{todonotes}  
\usepackage[normalem]{ulem} 
\usepackage{soul} 

\usepackage{algorithm,algorithmicx,algpseudocode}

\algnewcommand\algorithmicinput{\textbf{Input:}}
\algnewcommand\Input{\item[\algorithmicinput]}

\algnewcommand\algorithmicoutput{\textbf{Output:}}
\algnewcommand\Output{\item[\algorithmicoutput]} 

\graphicspath{{figures/pdf/}}

\newbox\tempbox
\makeatletter
\let\NAT@parse\undefined
\makeatother

\usepackage{hyperref} 
\hypersetup{
  colorlinks   = true, 
  urlcolor     = blue, 
  linkcolor    = blue, 
  citecolor   = green 
}

\usepackage{balance}
\title{\LARGE \bf
Model-Based Pose Estimation of Steerable Catheters\\ under Bi-Plane Image Feedback}
\author{Jared Lawson$^{1}$, Rohan Chitale$^{2}$ and Nabil~Simaan$^{1}$$^{\dag}$
\thanks{$\dag$ Corresponding author}
\thanks{$^{1}$Department of Mechanical Engineering, Vanderbilt University, Nashville, TN 37235, USA
        {\tt\small (jared.p.lawson, nabil.simaan) @vanderbilt.edu}}%
\thanks{$^{2}$Department of Neurological Surgery, Vanderbilt University Medical Center, Nashville, TN 37235, USA
        {\tt\small (rohan.chitale) @vumc.org}}%
\thanks{J. Lawson was supported by NIH award \#T32EB021937 and by Vanderbilt University funds. The authors thank Madison Veliky for supporting experiments.}%
}

\begin{document}
\maketitle

\thispagestyle{empty}  


\thispagestyle{fancy}
\fancyhf{}
\renewcommand{\headrulewidth}{0pt}
\lhead{2023 IEEE International Conference on Robotics and Automation (ICRA). Accepted Version. }
\rfoot{\centering \scriptsize \copyright 2023 IEEE. Personal use of this material is permitted. Permission from IEEE must be obtained for all other uses, in any current or future media, including reprinting/republishing this material for advertising or promotional purposes, creating new collective works, for resale or redistribution to servers or lists, or reuse of any copyrighted component of this work in other works.}

\pagestyle{empty}

\begin{abstract}
Small catheters undergo significant torsional deflections during endovascular interventions. A key challenge in enabling robot control of these catheters is the estimation of their bending planes. This paper considers approaches for estimating these bending planes based on bi-plane image feedback. The proposed approaches attempt to minimize error between either the direct (position-based) or instantaneous (velocity-based) kinematics with the reconstructed kinematics from bi-plane image feedback.
A comparison between these methods is carried out on a setup using two cameras in lieu of a bi-plane fluoroscopy setup. The results show that the position-based approach is less susceptible to segmentation noise and works best when the segment is in a non-straight configuration. 
These results suggest that estimation of the bending planes can be accompanied with errors under $30^{\circ}$. Considering that the torsional buildup of these catheters can be more than $180^{\circ}$, we believe that this method can be used for catheter control with improved safety due to the reduction of this uncertainty.    
\end{abstract}

\section{Introduction} \label{sec:intro}
\par Interventional radiology (IR) enables the treatment of certain pathologies through an endovascular approach, thereby avoiding the trauma of open surgery. Interventional radiologists leverage different imaging modalities to visualize and guide endovascular navigation of catheters and guidewires to target anatomy. Neurointerventions represent a subset of IR procedures dedicated to treatment of vascular pathologies of the brain, including acute ischemic stroke relief by mechanical thrombectomy (MT) \cite{Berkhemer2015,Campbell2015,Goyal2015,Jovin2015,Saver2015}.
\par Highly tortuous vasculature complicates cerebral vessel navigation. Existing neurointerventional catheters are long (${\sim}1m$), passive and flexible, which further complicates the task of navigation. These challenges have led to research on steerable catheters which can control the bending of their distal tip using proximal actuation inputs.
\par An additional challenge affecting neuronavigation is the control of catheters under bi-plane imaging. Rather than having exact understanding of how the catheter is oriented in 3D space, the interventionalist must mentally reconstruct the separate planar images (an anterior-posterior (AP) view and a lateral view) to intuit how the catheter should be safely manipulated. This leads to significant cognitive burden and further complicates catheter navigation. 
\begin{figure*}[htbp]
        \centering
        \includegraphics[trim={0.7cm 0.6cm 0.5cm 0.3cm},clip, width=0.9\textwidth]{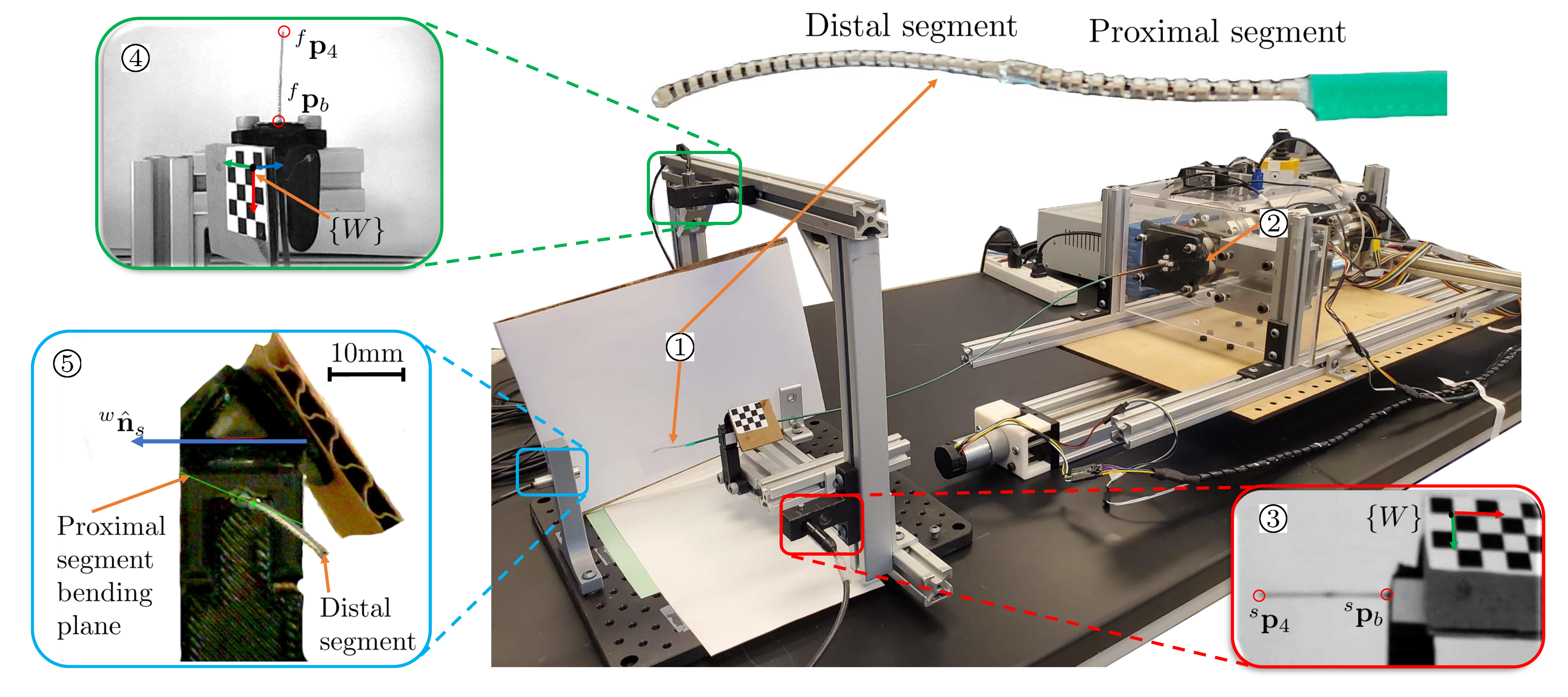}
        \caption{A two-segment bending catheter (top right \protect\circled{1}) showing the actuation unit (bottom right \protect\circled{2}). right camera view \protect\circled{3}, top camera view \protect\circled{4}, front camera view  \protect\circled{5}. }\label{fig:experimental_setup}
\end{figure*}
\par To address the challenges associated with steering microcatheters for neurointervention, a steerable robotic microcatheter (SRMC) similar to the one shown in Fig~\ref{fig:experimental_setup} \circled{1} was introduced in \cite{Abah2019,abah2021image}. This double-articulated microcatheter (0.89mm OD and 0.48mm ID) was designed specifically to traverse from the aortic arch into the internal carotid arteries (ICAs), and is robotically-actuated to achieve a desired bending configuration. One desired application for this system includes closed-loop control for semi-autonomous neuronavigation, which requires feedback of the catheter state from bi-plane fluoroscopic images, as initially demonstrated in \cite{Abah2019}. 
\par While bi-planar images provide a good estimate of the tip position and orientation, they provide no estimate of torsion, or roll, about the catheter's longitudinal direction. Catheters suffer significant torsional losses between the rotation input at their proximal end and the rotation output at their distal end. These torsional losses are due to their high slenderness ratio and the highly elastic catheter materials. Considering that torsional losses of these catheters can be more than $180^{\circ}$, these torsional effects prohibit effective control of catheters under bi-plane imaging. To achieve safe robotic control of such catheters, the angle of the natural plane of bending at the catheter's distal end must be known at all times. 
\par The contribution of this study explores a way to estimate these bending plane angles without reliance on mechanical modeling of torsional and frictional losses, but with few points segmented from bi-plane imaging. This assumes the distal and proximal ends of the bending segments of a steerable catheter can be equipped with at least two radio-opaque markers, which can be easily identified in bi-plane images. Using 3D position and velocity reconstruction, we compare the expected error in estimating the bending planes of these catheters if one were to use position or velocity information. The methods presented herein are valid for both constant and variable curvature segments including multiple stacked segments. 
\par In Section \ref{sec:related_works}, the previous works related to estimating catheter pose and shape are presented to motivate the contributions of this work. The robotic system and setup is described in Section \ref{sec:micro_cath_setup}. In Section \ref{sec:kinematics}, the direct and instantaneous kinematics are presented for use in the estimation approach which is described in detail in Section \ref{sec:estimationapproach}. Simulation and experimental studies of this approach are shown in Section \ref{sec:eval}, and conclusions and future works are synthesized in Section \ref{sec:conclusion}.
\section{Related Works} \label{sec:related_works}
%
%
\par Robotic catheter systems have been developed in both academic and industry settings for applications including structural vascular procedures and cardiac electrophysiology (EP), many of which are reviewed in \cite{Rafii-Tari2014,Hu2018_review,Zhao2022}. Most of the existing commercialized robotic catheter systems were developed with the sole aim of limiting radiation exposure to the interventional team. This is done by actuation units that allow remote telemanipulation from outside of the radiation field. While this addresses the occupational hazard associated with repeated radiation exposure for the interventionalist, it does not improve the distal dexterity of catheters or alleviate the cognitive burden of torsionally flexible catheter navigation using bi-plane imaging. 
\par Magnetic tracking was used for pose estimation in \cite{Jayender2008,Penning2012,Tercero2007}, but this approach comes at the cost of a significant cross-sectional real estate needed to accommodate the magnetic coils, complicates catheter sterilization, and increases costs. Other groups leveraged both 2D and 3D ultrasound imaging to estimate the pose of a catheter tip. Kesner et al. use 3D ultrasound to visualize both the catheter and heart wall for cardiac ablation \cite{Kesner2014}, while Boskma et al. use 2D ultrasound as a feedback modality for a magnetically-actuated catheter \cite{Boskma2016}. Optical fibers and fiber Bragg grating (FBG) sensors can estimate bending plane and catheter pose via curvature sensing, but remain expensive and sensitive to thermal effects \cite{Khan2020,AlAhmad2020}. Of the works reconstructing catheters in bi-plane fluoroscopic images, most cannot describe torsion about the catheter's axis \cite{bender1999,baert2003,baur2016,delmas2015,hoffmann2013,hoffmann2015,schenderlein2010,wagner20164d}. Papalazarou et al. consider this torsion but require shape information along the full bending segment \cite{papalazarou2012}. Single-plane fluoroscopy can be used to estimate the bending plane of a single bending segment if the segment is assumed to bend in a circular fashion. This approach requires segmenting the location of multiple radiopaque markers along the catheter \cite{Lim2020,Hwang2018}. A circular bending assumption is not a safe assumption for flexible catheters, in which multiple points of contact can be affecting the 3D shape of the catheter. Similarly, Camarillo et al. present a method to extract the shape of a bending manipulator by reconstructing point clouds of the manipulator's "silhouette" from multiple image planes \cite{Camarillo2008,Camarillo2009}, which can be useful for bi-plane fluoroscopy for guidewires, but may not be feasible for catheters that only have radiopaque marker bands at the tip and are radiolucent along their length. Ravigopal et al. utilize single C-arm images for reconstruction using deep learning techniques \cite{ravigopal2021,ravigopal2022}. Few papers have included fusion of sensing modalities for catheter pose or shape tracking \cite{Ha2021robust,Borgstadt2015multi}. Of these prior works, there still exists a need for pose estimation of steerable catheters that do not have full shape information in bi-plane fluoroscopic images, which cannot afford to utilize embedded sensors due to shape or design constraints.

\section{A two-Segment Steerable Micro Catheter} \label{sec:micro_cath_setup}
\par The SRMC robot shown in Fig.~\ref{fig:experimental_setup} \circled{1} is a two-segment continuum robot with two corresponding natural bending planes. Each segment uses push-pull antagonistic actuation of two superelastic NiTi tubes with eccentric flexures. The inner/outer diameter of the distal segment are 0.48/0.89mm, respectively. The proximal segment uses the outer tube of the distal segment as an inner tube and it is actuated by pushing/pulling on an outer tube with an outer diameter of 1.2mm. The proximal and distal segments are 13mm and 16mm, respectively. The robot is actuated by a 4 degree of freedom (DoF) actuation unit that includes insertion, rotation, and bending of each segment. Control is implemented on a Simulink Real-Time operating system with a PC104 (xPC) control computer. The control cycle frequency is set at 1KHz. 
\par Figure~\ref{fig:experimental_setup} also shows a mock bi-plane imaging setup that uses USB cameras \circled{3} and \circled{4} (640x480px, 20fps), in lieu of bi plane fluoroscopy images. A third camera \circled{5} giving a frontal view looking orthographically on the base of the proximal segment is also used for ground truth in determining the bending plane as discussed in Section \ref{sec:eval}. 


\begin{figure}[htbp]
        \centering
        \includegraphics[trim={0.2cm 1.9cm 0.2cm 0cm},clip,width=0.99\columnwidth]{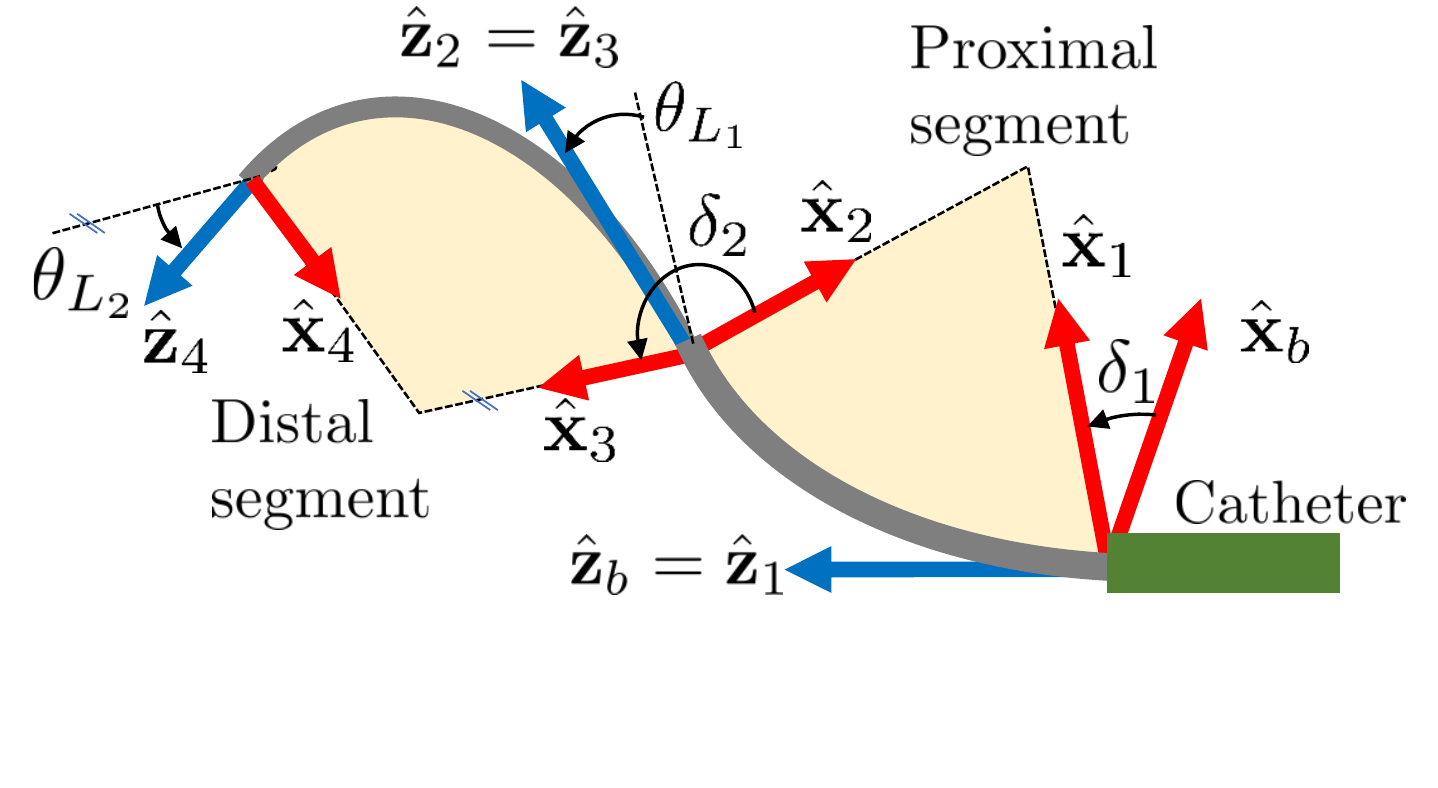}
        \caption{Steerable Catheter: Visualization of bending frames for double-articulating SRMC.}\label{fig:bending_frames}
\end{figure}
\section{Kinematics Formulation}\label{sec:kinematics}
\par The kinematics of the double-articulated steerable microcatheter has been previously introduced in \cite{Abah2019,abah2021image}. Unlike \cite{abah2021image}, we consider torsional losses along the length of the catheter and we do not include a passive flexure between the two bending segments of the steerable catheter.
\subsection{Direct Kinematics}\label{sec:dirkin}
We assign frame \{B\} as the base frame\footnote{The notation $\{A\}$ denotes a right-handed frame with origin at point $\mb{o}_a$ and unit vectors $\uvec{x}_a$, $\uvec{y}_a$, $\uvec{z}_a$.} at the base of the proximal segment with $\uvec{z}_b$ pointing distally along the axis of the catheter, $\uvec{x}_b$ aligned within the initial bending plane of the proximal segment, and $\uvec{y}_b$ completing a right-handed frame and denoting the sign of the bending angle according to the right-hand rule. For a given proximal actuation push/pull amount $q_p$ and at arc length location $s$, we define the local curve tangent angle $\theta_1(s,q_p)$ as the angle of the local tangent in plane $\uvec{x}_1$, $\uvec{z}_1$. We also use $\theta_{L_1}$ to denote the tip angle $\theta_1(s=L_1)$, where $L_1$ is the segment length. Finally, we define the complementary angle $\tilde{\theta}_{L_1} = \frac{\pi}{2}-\theta_{L_1}$. In the following derivations, we will use $\theta_1$ as a shorthand notation for $\theta_1(s,q_p)$.   
\par The first bending segment is assigned frame \{1\} as a simple rotation about $\uvec{z}_b$ by an angle $\delta_1$. Assuming that the actuation unit commands a base rotation angle $q_{r}$ and that a torsional transmission loss angle $\delta_{L}$ results from torsional friction buildup, the angle characterizing the bending plane of the proximal segment can be written $\delta_{1} = q_{r} + \delta_{L}$. The direct kinematics of the first bending segment becomes\footnote{The notation $\mb{a}^\wedge$ denotes the cross-product form of vector $\mb{a}$}:
\begin{subequations}\label{eq:dk1}
  \begin{equation}\label{eq:dkp1}
  ^{b}\mb{o}_{2/b} = 
  e^{\delta_{1}[\hat{\mb{z}}_{b}]^\wedge}
  \int_{0}^{L_1}\left[\cos(\theta_1), 0, \sin(\theta_1)
  \right]\T ds
  \end{equation}
  \begin{equation}\label{eq:dkR1}
    ^{b}\mb{R}_{2} = e^{\delta_{1}[\hat{\mb{z}}_{b}]^\wedge}
e^{\tilde{\theta}_{L_{1}}[\hat{\mb{y}}_{1}]^\wedge}                           
  \end{equation}
\end{subequations}
where $^{b}\mb{o}_{2/b}$ is the tip position of segment 1 relative to the origin of \{B\} and $^{b}\mb{R}_{2}$ is the orientation of frame \{2\} at the tip of the first segment\footnote{We use ${}^b\mb{o}_{a/b}$ to denote vector $\mb{o}_{a/b}$ represented in frame \{B\} and ${}^b\mb{R}_a$ to denote the orientation of frame \{A\} relative to frame \{B\}}. 

\par Using $q_d$, $\delta_2$, $L_2$, $\theta_2$ and $\tilde{\theta}_{L_2}$ in place of $q_p$, $\delta_1$, $L_1$, $\theta_1$ and $\tilde{\theta}_{L_1}$ the kinematics of the distal segment with respect to \{2\} follows the same derivation, returning $^{2}\mb{o}_{4/2}$ and $^{2}\mb{R}_{4}$. In this expression $q_d$ and $L_2$ denote the actuation and the length of the distal segment, and the angle $\theta_{L_2}$ is defined for the distal segment in a similar fashion as for the proximal segment. The angle $\delta_2$ defines the distal bending plane as rotated about $\uvec{z}_2$ according to the right-hand rule. Since the first bending segment is short (relative to catheter length), we neglect its contribution to torsional loss. Hence, $\delta_2$ is assumed fixed. Finally, $\theta_2$ is used as a shorthand notation for $\theta_2(s,q_d)$.   
%
%
\par Given the kinematics of each independent segment, the full configuration of the catheter with respect to $\{B\}$ is written:
\begin{subequations} \label{eq:DKfull}
\begin{equation} \label{eq:DKfulla}
    ^{b}\mb{o}_{4/b} = \leftidx{^b}{\mb{o}_{2/b}} + \leftidx{^b}{\mb{R}_{2}}\leftidx{^2}{\mb{o}_{4/2}}
\end{equation}
\begin{equation}
    ^{b}\mb{R}_{4} = \leftidx{^b}{\mb{R}_{2}}\leftidx{^2}{\mb{R}_{4}}
\end{equation}
\end{subequations}
\par Since the segments do not necessarily bend in a circular fashion, we characterize the bending shape using the local tangent angle to the backbone of each segment ($\theta_i(s,q)$ where $i=1,2$). The change in bending curvature profile as a function of actuation can be captured using a modal representation yielding a double interpolation least-squares solution that best fits experimental calibration data as in \cite{abah2021image}. This process results in the modal kinematics for both segments given by:
\begin{equation}\label{eq:modal}
    \theta_i(s,q_{j}) = \bs{\psi}(s)^{T}\mb{A}_{i}\bs{\eta}(q_{j})
\end{equation}
where ${j=p,d}$, $\bs{\psi}\in\realfield{n}$ represents the vector of modal basis functions, $\bs{\eta}\in\realfield{m}$ the vector of modal coefficients, and $\mb{A}_{i}\in\realfield{n\times m}$, $i=1,2$, denote the characteristic shape matrices obtained from a least-squares calibration as in \cite{abah2021image}.
\subsection{Instantaneous Kinematics}
\par The instantaneous direct kinematics is obtained via differentiation of the direct kinematics. The linear and angular velocity\footnote{we use ${}^b\bs{\omega}_a$ to denote the angular velocity vector of frame \{A\} described in frame \{B\}} of the tip of the first segment are given by:
\begin{subequations}\label{eq:IK1}
  \begin{equation}\label{eq:IKv1}
    ^{b}\mb{v}_{2} = \begin{bmatrix}
                       -\dot{\delta}_{1}\sin(\delta_{1}) \nu-\cos(\delta_{1})\int_{0}^{L_1}\sin(\theta_1)\dot{\theta}_{1}ds \\
                       \dot{\delta}_{1}\cos(\delta_{1}) \nu-\sin(\delta_{1})\int_{0}^{L_1}\sin(\theta_1)\dot{\theta}_{1}ds \\
                       \int_{0}^{L_1}\cos(\theta_1)\dot{\theta}_{1}ds
                     \end{bmatrix}
  \end{equation}
  \begin{equation}\label{eq:IKw1}
    ^{b}\bs{\omega}_{2} = \left[\dot{\theta}_{1}\sin\delta_{1},~
                            -\dot{\theta}_{1}\cos\delta_{1},~
                            \dot{\delta}_{1}
                          \right]\T
  \end{equation}
\end{subequations}
where $\nu$ is the horizontal coordinate along $\uvec{x}_1$:
\begin{equation}\label{eq:dx}
    \nu = \int_{0}^{L_1}\cos(\theta_1)ds
\end{equation}
The integral terms in the first two rows of Equation \eqref{eq:IKv1} include $\dot{\theta}_1$, which is a function of both arc-length and time. This term can be further reduced to relate the velocity at the tip of the segment to the velocity of the joint variable, $\dot{q}_p$:
\begin{equation}
    \int_{0}^{L_1}\sin(\theta_1)\dot{\theta}_{1}ds = \dot{q}_{p}\int_{0}^{L_1}\sin(\theta_1)\frac{\partial \theta_{1}}{\partial \bs{\eta}}\frac{\partial \bs{\eta}}{\partial q_{p}}ds
\end{equation}
where $\frac{\partial \theta_{1}}{\partial \eta}$ is found by taking the partial derivative of \eqref{eq:modal}a:
\begin{equation}
    \frac{\partial \theta_{1}}{\partial \bs{\eta}} = \bs{\psi}(s)^{T}\mb{A}_{1}
\end{equation}
\par Using the above definitions, we derive the instantaneous direct kinematics geometric Jacobian $\mb{J}_1\in\realfield{6\times 2}$ such that $\bs{\xi}_1=\mb{J}_1[\dot{q_{p}},\dot{\delta}_1]\T$ where $\bs{\xi}_1$ is the twist of frame \{2\} with linear velocity preceding angular velocity. The first 3 rows of $\mb{J}_1$ denote the translational Jacobian $\mb{J}_{v1}$ and the last 3 rows the orientational Jacobian $\mb{J}_{\omega 1}$ given by: 
\begin{subequations}
 \begin{equation}\label{eq:Jv1}
    \mb{J}_{v1} = \begin{bmatrix}
      -\cos\delta_{1}\int_{0}^{L_1}\sin(\theta_1)\bs{\psi}(s)^{T}\mb{A}_{1}\frac{\partial \bs{\eta}}{\partial q_{p}}ds & -\sin\delta_{1} \nu \\
      -\sin\delta_{1}\int_{0}^{L_1}\sin(\theta_1)\bs{\psi}(s)^{T}\mb{A}_{1}\frac{\partial \bs{\eta}}{\partial q_{p}}ds & \cos\delta_{1} \nu \\
      \int_{0}^{L_1}\cos(\theta_1)\bs{\psi}(s)^{T}\mb{A}_{1}\frac{\partial \bs{\eta}}{\partial q_{p}}ds & 0
    \end{bmatrix}
\end{equation}
\begin{equation}\label{eq:JOmega1}
    \mb{J}_{\omega 1} = \begin{bmatrix}
      \bs{\psi}(s)^{T}\mb{A}_{1}\frac{\partial \bs{\eta}}{\partial q_{p}}\sin\delta_{1} & 0 \\
      -\bs{\psi}(s)^{T}\mb{A}_{1}\frac{\partial \bs{\eta}}{\partial q_{p}}\cos\delta_{1} & 0 \\
      0 & 1
    \end{bmatrix}
\end{equation}
\end{subequations}
%
%
In this equation, a simplifying assumption is that the rate of torsional loss $\dot{\delta}_{L}$ is negligible, such that the rate of change of the bending plane $\dot{\delta}_{1}\approx\dot{q}_{r}$. This assumption is acceptable since the segmentation and estimation protocol can run at a fast rate of around 20Hz.
\par The instantaneous direct kinematics of the distal segment with respect to \{2\} follows the same derivation and yields the Jacobian $\mb{J}_2$ such that $\bs{\xi}_2=\mb{J}_2\left[\dot{q}_d,\dot{\delta}_2\right]\T$
where $\dot{\delta}_2=0$ is used since the distal segment is affixed to the proximal segment. This Jacobian follows the same expression for its translational and rotational parts as in \eqref{eq:Jv1} and \eqref{eq:JOmega1}, but replaces $\theta_1$, $\delta_1$, $q_p$, and $\mb{A}_1$ with $\theta_2$, $\delta_2$, $q_d$, and $\mb{A}_2$. Using these segment Jacobians, the combined serial kinematics becomes:
\begin{equation} \label{eq:IKfinal}
    \bs{\xi}\triangleq\begin{bmatrix}
      ^{b}\mb{v}_{4} \\
      ^{b}\bs{\omega}_{4}
    \end{bmatrix} = \begin{bmatrix}
      \mb{S}_{1}\mb{J}_{1} &| & \mb{S}_{2}\mb{J}_{2}
    \end{bmatrix} \begin{bmatrix}
      \dot{q}_{p} \\
      \dot{q}_{r} \\
      \dot{q}_{d} \\
      0
    \end{bmatrix}
\end{equation}
where $\mb{S}_{1}$ and $\mb{S}_{2}$ represent the twist transformation matrices (adjoints) given by:
\begin{equation}
    \mb{S}_{1} = \begin{bmatrix}
      \mb{I} & \left(-{^{b}\mb{R}_{2}}{^{2}\mb{o}_{4/2}}\right)^{\wedge} \\
      \mb{0} & \mb{I}
    \end{bmatrix},
\quad 
    \mb{S}_{2} = \begin{bmatrix}
      \leftidx{^b}{\mb{R}_{2}} & \mb{0} \\
      \mb{0} & \leftidx{^b}{\mb{R}_{2}}
    \end{bmatrix}
\end{equation}
where $\mb{I}\in\realfield{3\times 3}$ is the identity matrix. 
\begin{figure}[htbp]
        \centering
        \includegraphics[width=0.8\columnwidth]{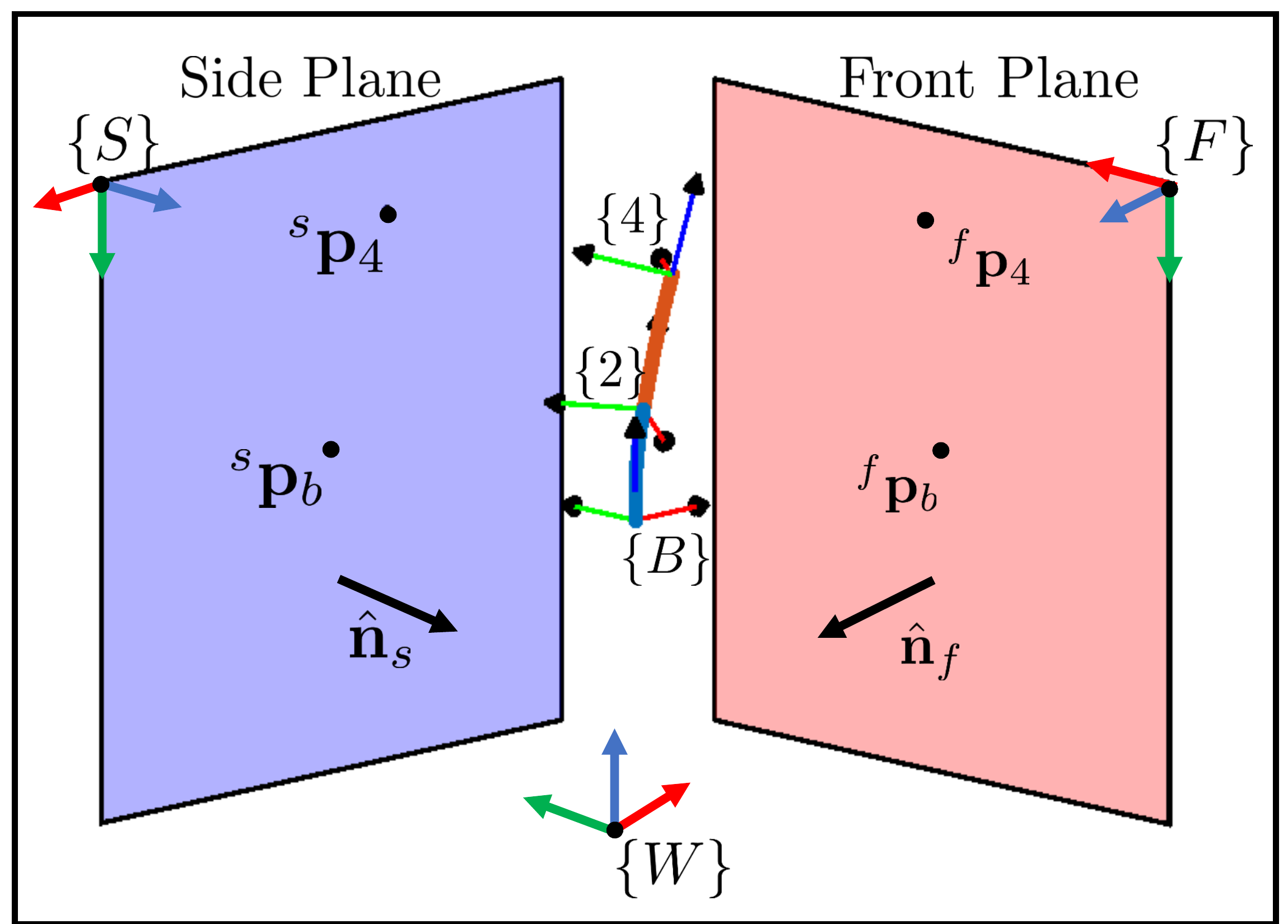}
        \caption{Bi-Plane Projections of Catheter Frames. }\label{fig:proj_planes}
\end{figure}
\begin{figure*}[htbp]
        \centering
        \includegraphics[width=0.9\textwidth]{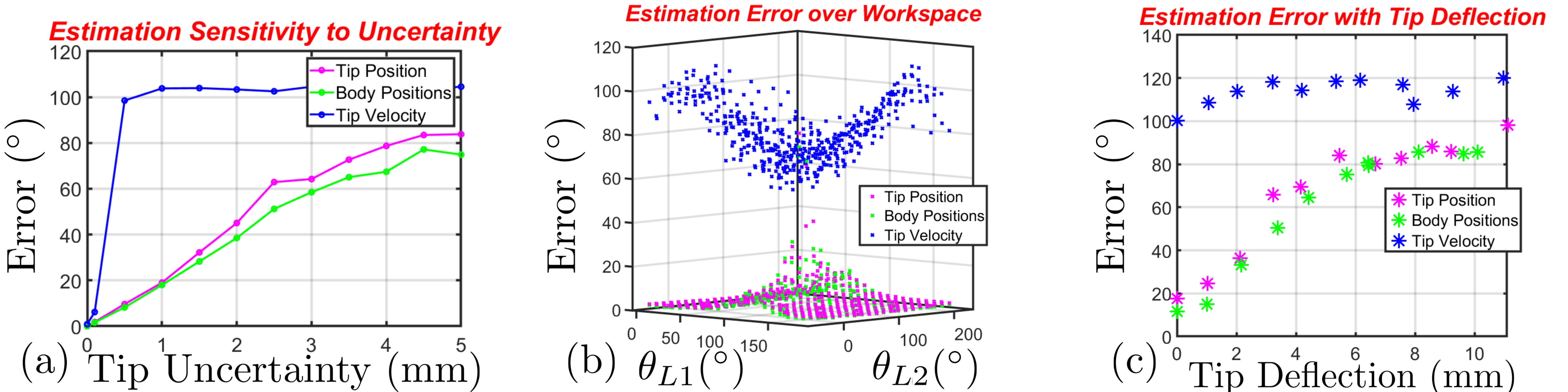}
        \caption{Sensitivity Analysis: (a) Estimation error as noise is added to the image projections. (b) Configuration-dependent estimation error over the workspace assuming 1mm of segmentation noise. (c) Estimation error subject to varying levels of tip deflection (in the bending direction) with the environment, also assuming 1mm of segmentation noise.}\label{fig:sensitivity_plots}
\end{figure*}
\section{Bending Plane and Pose Estimation Approach} \label{sec:estimationapproach}
\par The image planes in neurointervention are the anterior-posterior (AP) and lateral views, which are respectively labeled the front $\{F\}$ and side $\{S\}$ planes in Figure \ref{fig:proj_planes} and the remainder of this paper.  
The orientation of these imaging planes is defined by their respective unit normal vectors, $\hat{\mb{n}}_{f}$ and $\hat{\mb{n}}_{s}$, which are known in the world coordinate frame $\{W\}$ from the encoders on the bi-plane fluoroscopy machine.
\par In both  planes, image segmentation algorithms can extract the positions of discrete points along the length of the catheter where radiopaque marker bands are placed \cite{VanAssan2000,Hwang2018}. The minimal required input for the proposed approach will assume segmentation at both the tip and base of the catheter's bending segment, corresponding to the coordinate frames labeled $\{4\}$ and $\{B\}$ in Figure \ref{fig:proj_planes}, respectively. The segmented positions of these frames in the local frame of the front plane will be denoted ${}^{f}\tilde{\mb{o}}_{b}$ and ${}^{f}\tilde{\mb{o}}_{4}$. Similarly, ${}^{s}\tilde{\mb{o}}_{b}$ and  ${}^{s}\tilde{\mb{o}}_{4}$ denote the segmented positions of these frames in the local frame of the side plane. 
\par The vector ${}^w\mb{o}_{4/b}\triangleq{}^w\mb{o}_4-{}^w\mb{o}_b$ represents the tip location relative to the base. This vector is projected onto the front and side planes via these projection matrices \cite{Basilevsky2013}:
\begin{equation}\label{eq:projmat}
\leftidx{^w}{\mb{P}}_{f} = \mb{I} - \hat{\mb{n}}_{f} \hat{\mb{n}}_{f}^{T}, \qquad
\leftidx{^w}{\mb{P}}_{s} = \mb{I} - \hat{\mb{n}}_{s} \hat{\mb{n}}_{s}^{T}
\end{equation} 
The projection of ${}^w\mb{o}_{4/b}$ onto \{F\} and \{S\} is given by:
\begin{equation}
{}^w\mb{P}_f {}^w\mb{o}_{4/b}={}^w\mb{R}_f{}^f\tilde{\mb{o}}_{4/b}, \quad
{}^w\mb{P}_s {}^w\mb{o}_{4/b}={}^w\mb{R}_s{}^s\tilde{\mb{o}}_{4/b}
\end{equation}
where ${}^f\tilde{\mb{o}}_{4/b}\triangleq\left({}^f\tilde{\mb{o}}_4-{}^f\tilde{\mb{o}}_b\right)$ and ${}^s\tilde{\mb{o}}_{4/b}\triangleq\left({}^s\tilde{\mb{o}}_4-{}^s\tilde{\mb{o}}_b\right)$ represent the difference between the segmented coordinate frames in the front and side imaging planes. 
\par Given the above equations, one can estimate ${}^w\mb{o}_{4/b}$ to find the reconstructed position of the catheter tip relative to the base of the first segment in world frame:
%
\begin{equation} \label{eq:estDK}
    \leftidx{^w}{\tilde{\mb{o}}}_{4/b} = \begin{bmatrix}
    \leftidx{^w}{\mb{P}}_{f} \\
    \leftidx{^w}{\mb{P}}_{s}
    \end{bmatrix}^{+} \begin{bmatrix}
    \leftidx{^w}{\mb{R}}_{f}\left(\leftidx{^f}{\tilde{\mb{o}}}_{4} - \leftidx{^f}{\tilde{\mb{o}}}_{b}\right) \\
    \leftidx{^w}{\mb{R}}_{s}\left(\leftidx{^s}{\tilde{\mb{o}}}_{4} - \leftidx{^s}{\tilde{\mb{p}}}_{b}\right)
    \end{bmatrix}
\end{equation}
\par The bending plane can be estimated by finding the loss parameter, $\delta_{L}$, which minimizes the difference between the modeled direct kinematics \eqref{eq:DKfulla} and the estimated direct kinematics \eqref{eq:estDK}. Since this error is a nonlinear function of $\delta_{L}$, the following nonlinear least-squares statement defines the problem:
%
\begin{equation} \label{eq:solveang1}
    \delta_{L}^{*} = \argmin\left(\left(\textrm{DirKin}\left(\mb{q},\,\delta_{L}\right) - \leftidx{^w}{\tilde{\mb{o}}}_{4/b}\right)^{2}\right)
\end{equation}
where $\textrm{DirKin}\left(\mb{q},\,\delta_{L}\right)$ refers to the direct kinematics as described in \eqref{eq:dk1}-\eqref{eq:DKfull}, which is a function of the joint value vector $\mb{q}\triangleq\left[q_r , q_p , q_d\right]\T$ and the torsional loss angle.
\par If more discrete points along the robot are available from segmentation, \eqref{eq:solveang1} can be extended to compare the model and estimated kinematics for the set of discrete points:
\begin{equation} \label{eq:solveang2}
    \delta_{L}^{*} = \argmin\left(\sum_{i=1}^{n}\left(\textrm{DirKin}_{i}\left(\mb{q},\,\delta_{L}\right) - \leftidx{^w}{\tilde{\mb{o}}}_{i/b}\right)^{2}\right)
\end{equation}
where $n$ is the total number of segmented points and $\textrm{DirKin}_{i}$ is the direct kinematics up to the $i^{th}$ point.
\par An alternate approach that is considered leverages the use of the instantaneous kinematics to find the bending plane angle based on the displacement of the catheter tip over time. The direction of the tip velocity is found by finite differences of segmented tip positions, and will be noted as $\left(\leftidx{^w}{\dot{\tilde{\mb{o}}}}_{4/b}\right)$. \eqref{eq:solveang1} and \eqref{eq:solveang2} can be further modified to compare the squared error of the instantaneous kinematics:
\begin{equation} \label{eq:solveang3}
    \delta_{L}^{*} = \argmin\left(\left(\textrm{InstKin}\left(\dot{\mb{q}},\,\delta_{L}\right) - \leftidx{^w}{\dot{\tilde{\mb{o}}}}_{4/b}\right)^{2}\right)
\end{equation}
where $\textrm{InstKin}\left(\dot{\mb{q}},\,\delta_{L}\right)$ refers to the instantaneous kinematics as described in \eqref{eq:IKfinal}, which are a function of the joint speed vector $\dot{\mb{q}}$ and the torsional loss angle.
\begin{figure*}[htbp]
        \centering
        \includegraphics[width=0.9\textwidth]{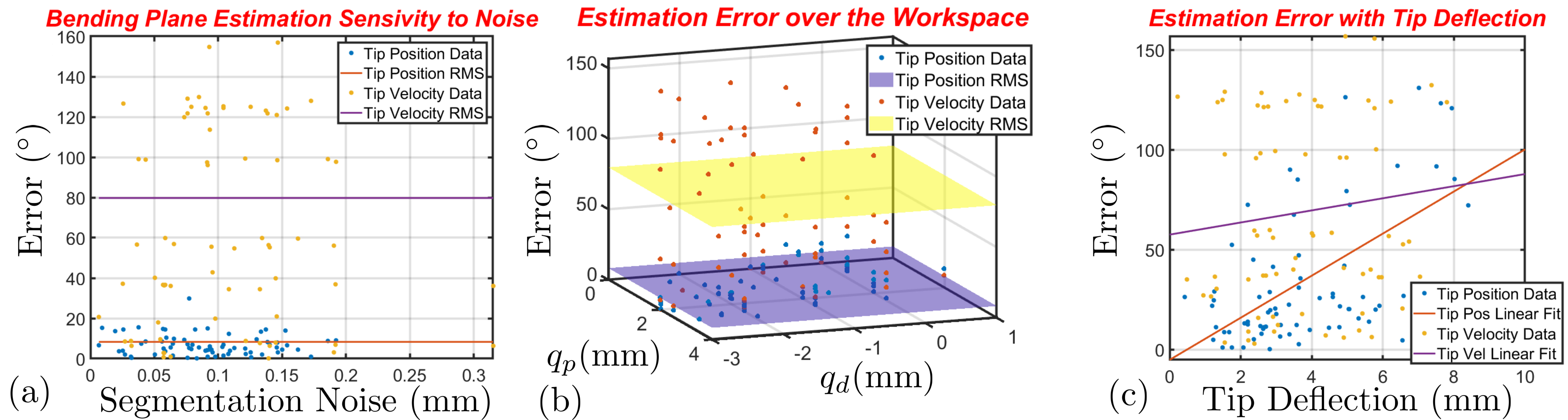}
        \caption{Experimental Results: (a) Estimated bending angle error across recorded segmentation noise levels. (b) Estimation error over the workspace showing sensitivity to increased distal bending. (c) Increasing estimation error with higher levels of tip deflection from experimental data.}\label{fig:experimental_results}
\end{figure*}
%
\section{Evaluation} \label{sec:eval}
\subsection{Simulation Study: Sensitivity Analysis}
\par In Section \ref{sec:estimationapproach}, three estimation approaches were presented. In this section, the approach outlined in \eqref{eq:solveang1} will be annotated as the "Tip Position" approach, while the approaches in \eqref{eq:solveang2} \& \eqref{eq:solveang3} will be labeled "Body Positions" and "Tip Velocity" approaches, respectively. The different approaches were simulated in MATLAB using lsqnonlin() nonlinear least-squares optimization with initial guesses of $0^{\circ}$. The approaches were simulated to compare their sensitivity to the expected disturbances that may arise in a clinical setting.
\par Fluoroscopic segmentation methods have been shown to exhibit inaccuracies ${<5}mm$ due to low signal-to-noise (SNR) ratio and segmentation algorithm sensitivity \cite{Wu2013,Wagner2016,Pauly2010,Chang2016,Ambrosini2017}. The segmentation approach developed from \cite{Abah2019} exhibits error ${<2}mm$. To compare the proposed estimation approaches' robustness to segmentation uncertainty, each method was simulated 500 times in the same configuration with different levels of random error projected on the coordinates in the imaging planes to simulate noisy segmented poses. Figure \ref{fig:sensitivity_plots}(a) shows that the bending plane estimation error increases with increasing levels of segmentation uncertainty. Overall, the "Body Positions" approach returns slightly improved error compared to the "Tip Position" approach, while the "Tip Velocity" approach is most sensitive to noise.
\par The bending configuration of the steerable catheter tip affects the utility of the estimation methods. For example, in straight configurations (no bending exhibited), there is no bending plane to solve for. The estimation error, assuming 1mm of segmentation uncertainty, was simulated for each approach across the catheter workspace, as shown in Figure \ref{fig:sensitivity_plots}(b). For the position-based approaches the error clearly increases approaching the straight configuration $\left(\theta_{L1}\rightarrow 90^{\circ},\,\theta_{L2}\rightarrow 90^{\circ}\right)$. Across a majority of the workspace, however, the position-based approaches are shown to have significantly low error subject to only 1mm of segmentation noise.
\par The catheter bending segments may be subject to deflection as they contact the anatomy, so this estimation approach must be robust to such a disturbance. Figure \ref{fig:sensitivity_plots}(c) shows the effect of tip deflections on the estimation approaches, assuming 1mm of segmentation uncertainty. It is important to specify that tip deflection here occurs in the bending plane direction. The results show a similar trend to the sensitivity to noise found in Figure \ref{fig:sensitivity_plots}(a).
\subsection{Experimental Evaluation}
\par The experimental setup is described in Section \ref{sec:micro_cath_setup} and shown in Fig~\ref{fig:experimental_setup} (and the multimedia extension).
\subsubsection{Protocol}
\par Mounted to the catheter-supporting fixture is a checkerboard fiducial marker for USB camera calibration which is visualized in both camera frames (shown in Fig~\ref{fig:experimental_setup} \circled{3} \circled{4}). Using MATLAB's built in Camera Calibrator App, which utilizes Zhang's planar fiducial camera calibration algorithm \cite{Zhang2000}, the pose of the fiducial marker origin frame \{W\} is found with respect to both USB camera frames, \{F\},\{S\}. Given the pose of each camera frame with respect to the assigned world frame (fiducial origin), the image plane unit normals are given by the z-axis direction of the camera planes with respect to the world $\left(\leftidx{^w}{\hat{\mb{n}}_{f}}=\leftidx{^w}{\hat{\mb{z}}_{f}},\,\leftidx{^w}{\hat{\mb{n}}_{s}}=\leftidx{^w}{\hat{\mb{z}}_{s}}\right)$.
\par Given the calibrated bi-plane imaging setup, the catheter is segmented in both image planes independently following an online variation of the approach first presented in \cite{Abah2019}. The segmentation approach subtracts the catheter from an a-priori background image to extract the tip of the catheter, and the base of the bending segments is fixed relative to the fiducial frame \{W\}, and runs at ${}\sim20$Hz. Segmentation of intermediate points along the catheter is feasible for fluoroscopic images, but infeasible for camera images, hence the Body Positions approach could not be validated, but will later be implemented for fluoroscopic use. 
%
\par With the catheter in a straight configuration, the catheter was rotated to an arbitrary bending plane. With the segmentation script running, the catheter was actuated to different joint configurations, pausing at each configuration to record ${}\sim100$ reconstructed tip positions and to record the ground truth image. Once the catheter returned to it's straight configuration, the actuation unit was rotated to a different bending plane, and the procedure was repeated at bending planes around the catheter's longitudinal axis.  
\par With knowledge of the joint positions returned from the actuation unit at each configuration, as well as the stream of segmented tip positions, the Tip Position and Tip Velocity estimation approaches described in \eqref{eq:solveang1}\&\eqref{eq:solveang3} were used to estimate each bending plane rotation. The ground truth bending plane was found by manually segmenting the proximal tip \{2\} of the steerable catheter as it was actuated in a single bending plane, which returns a vector of points that can be fit to a line representing the bending plane, shown as the green line in Fig. \ref{fig:experimental_setup} \circled{5}.
\par To determine the effect of tip deflection on these estimation approaches, the same segmented positions were used, however random noise was artificially added to the joint level encoder readings to deflect the modeled kinematics away from the estimated kinematics. 
\subsubsection{Results}
\par The experimental protocol was performed at 11 different bending planes, with a total of 94 joint configurations estimated across these 11 bending planes. Including the estimation of all bending plane configurations, the overall root-mean-squared (RMS) error between the estimated bending plane orientation and the ground truth orientation was found to be $8.31^{\circ}$ for the Tip Position approach, and the maximum estimation error was found to be $29.94^{\circ}$. The Tip Velocity approach exhibited RMS and maximum errors of $79.74^{\circ}$ and $156.8^{\circ}$, respectively. Across the workspace, with results shown in Figure \ref{fig:experimental_results}(b), there is no trend for errors of the Tip Velocity approach, however it can be observed that the Tip Position errors increase towards the straight configuration, as observed in Figure \ref{fig:sensitivity_plots}(b).
\par The segmentation noise at each of the configurations was characterized by the Euclidean norm of the standard deviation of each of the three segmented position components. The RMS segmentation noise across all data collected, shown in Fig. \ref{fig:experimental_results}(a), was found to be 0.11mm, while the maximum segmentation noise recorded for a particular configuration was found to be 0.35mm. There isn't a noticeable trend on the effect of segmentation noise on estimation error, but this is likely due to the relatively low noise exhibited from segmentation with this setup. 
\par By adding random joint level noise, tip deflection of up to 10mm was introduced into the experimental data. In Figure \ref{fig:experimental_results}(c), the estimation error increases as the tip deflection increases, which matches the simulation case from Figure \ref{fig:sensitivity_plots}(c). 
\section{Conclusion}\label{sec:conclusion}
\par In this study, methods to estimate the bending plane and pose of a steerable catheter tip are presented leveraging the catheter tip's calibrated kinematic model and bi-plane image feedback. Sensitivity analyses simulated the effects of bi-plane imaging noise, degree of bending, and tip deflection on the estimation accuracy of the presented methods. Experimental results validated the feasibility of this approach, with the tip position-based estimation approach returning the bending plane with an average accuracy around $8^{\circ}$, which greatly improves the rotation uncertainty observed for catheters which can be higher than $180^{\circ}$. Position-based estimation approaches are significantly more robust to segmentation uncertainty compared with velocity-based approaches throughout a majority of the catheter's workspace.
\par This study was limited to using grayscale USB camera images, rather than fluoroscopic X-ray images which are currently used in the clinical setting. Future studies will include implementing these estimation approaches under fluoroscopic imaging with added filtering, and will look at leveraging this estimation for feedback control of the catheter pose in phantom anatomies.

\bibliographystyle{IEEEtran}
\bibliography{bib/jared_bib,bib/nabil_bib}

\balance
\end{document}